\pdfoutput=1
\documentclass[a4paper]{article}

\usepackage{graphicx}
\usepackage{pifont,latexsym,ifthen,theorem,rotating,calc,textcase,booktabs,color}
\usepackage{amsfonts,amssymb,amsbsy,amsmath}
\usepackage{moreverb}
\usepackage[colorlinks,bookmarksopen,bookmarksnumbered,citecolor=black,urlcolor=black,linkcolor=black]{hyperref}
\usepackage[utf8]{inputenc}
\usepackage{dsfont} 
\usepackage[section]{placeins}
\usepackage{cite}
\usepackage[left=2.5cm,right=2.5cm,top=2cm,bottom=3cm,includeheadfoot]{geometry}
\usepackage{caption} 
\usepackage{mdframed,xcolor}
\newcommand{\fe}[1]{\mbox{\boldmath$#1$}}

\begin{document}

\title{Unbiased split variable selection for random survival forests using maximally selected rank statistics}

\author{Marvin~N.~Wright$^1$, Theresa~Dankowski$^1$ and Andreas~Ziegler$^{1-4}$}

\date{Dec. 2016}

\maketitle

\begin{mdframed}[backgroundcolor=blue!20] 
\noindent This is the peer reviewed version of the following article: \\[1mm]
Wright, M. N., Dankowski, T. \& Ziegler, A. (2017). Unbiased split variable selection for random survival forests using maximally selected rank statistics. Statistics in Medicine 36:1272–1284, \\[1mm]
which has been published in final form at \url{http://dx.doi.org/10.1002/sim.7212}. This article may be used for non-commercial purposes in accordance with Wiley Terms and Conditions for Use of Self-Archived Versions.
\end{mdframed}

\begin{abstract}
The most popular approach for analyzing survival data is the Cox regression model. The Cox model may, however, be misspecified, and its proportionality assumption may not always be fulfilled. An alternative approach for survival prediction is random forests for survival outcomes. The standard split criterion for random survival forests is the log-rank test statistics, which favors splitting variables with many possible split points. Conditional inference forests avoid this split variable selection bias. However, linear rank statistics are utilized by default in conditional inference forests to select the optimal splitting variable, which cannot detect non-linear effects in the independent variables. An alternative is to use maximally selected rank statistics for the split point selection. As in conditional inference forests, splitting variables are compared on the p-value scale. However, instead of the conditional Monte-Carlo approach used in conditional inference forests, p-value approximations are employed. We describe several p-value approximations and the implementation of the proposed random forest approach. A simulation study demonstrates that unbiased split variable selection is possible. However, there is a trade-off between unbiased split variable selection and runtime. In benchmark studies of prediction performance on simulated and real datasets the new method performs better than random survival forests if informative dichotomous variables are combined with uninformative variables with more categories and better than conditional inference forests if non-linear covariate effects are included. In a runtime comparison the method proves to be computationally faster than both alternatives, if a simple p-value approximation is used.
\end{abstract}

\footnotetext[1]{Institut f{\"u}r Medizinische Biometrie und Statistik, Universit{\"a}t zu L{\"u}beck, Universit{\"a}tsklinikum Schleswig-Holstein, Campus L{\"u}beck, L{\"u}beck, Germany}
\footnotetext[2]{Zentrum f{\"u}r Klinische Studien, Universit{\"a}t zu L{\"u}beck, L{\"u}beck, Germany}
\footnotetext[3]{School of Mathematics, Statistics and Computer Science, University of KwaZulu-Natal, Pietermaritzburg, South Africa}
\footnotetext[4]{Deutsches Zentrum f{\"u}r Herz-Kreislauf-Forschung, Standort Hamburg/Kiel/L{\"u}beck, L{\"u}beck, Germany}

\section{Introduction}\label{sec:introduction}
Survival data are generally analyzed using regression models, such as the Cox proportional hazards model \cite{Cox1972}. The standard Cox model is not well suited for big data problems, such as large scale omics studies or image analysis. Regularized Cox models may be used in this case \cite{Tibshirani1997}. However, the model may be misspecified, and the assumption of proportional hazards may also be violated. A promising alternative for survival prediction is the use of machine learning methods, such as survival trees or random forests \cite{Breiman2001}, and several tree-based approaches have been proposed for survival outcome \cite{Davis1989, Hothorn2004, Keles2002, Bou-Hamad2011}. The two major algorithms for random forests with survival outcome are random survival forests \cite[RSF]{Ishwaran2008} and conditional inference forests \cite[CIF]{Hothorn2006}. Recently, the statistical properties of random forests, including random survival forests have been better understood. Specifically, results have been obtained on consistency, convergence rate and asymptotic normality of random forest estimators \cite{Scornet2015,Ishwaran2010a,Wager2015,Wager2014b}.

RSF are closely related to the original random forests approach by Breiman \cite{Breiman2001}. Specifically, at each node the best split is selected as the maximum of the log rank statistic over all possible split points over all independent variables available for splitting at that node. In the standard R package for RSF \texttt{randomForestSRC} \cite{Ishwaran2014}, tree growing is stopped if the terminal node size is below a pre-defined threshold. A disadvantage of this random forests approach is that it favors splits for covariates with many possible split points \cite{Boulesteix2015, Loh2014, Strobl2007, Ziegler2014}. For illustration, consider a dataset with response variable $y$ and two covariates $x_1$ and $x_2$ with $n_1$ and $n_2$ possible cut points, respectively. Further, suppose that $y$ is independent of both $x_1$ and $x_2$, and $n_1 < n_2$. If we now search for the best split point by comparing all possible split points of both independent variables for their effect on $y$, $x_2$ will have a higher probability to have the split point with the larger effect on $y$---just by chance. If the split variable selection is biased, other parameter estimates, such as variable importance measures are biased as well \cite{Strobl2007}. Predictions could also suffer from the underestimation of important variables with few categories. In particular, if gene expression data and single nucleotide polymorphisms (SNP), both measured on microarrays, are to be analyzed jointly, the quantitative gene expression levels will be generally preferred for splitting by the RSF algorithm over the discrete SNP data. In fact, large-scale microarray data contain many uninformative, i.e., noise variables, and they are potentially mixed with important clinical covariables. This increases the need for unbiased split variable selection \cite{Yang2010}.

An approach to avoid split variable selection bias are CIF. CIF are conceptually different from standard random forests because they separate the selection of the split variable from selection of the split point of the already selected split variable \cite{Hothorn2006}. In the first step, the optimal split variable is determined. Specifically, an association test is performed between the possible split variables and the response. In the \texttt{party} package \cite{Hothorn2014}, this association test is by default a linear rank test. For survival data, a log-rank transformation for censored data is performed. If the association is found to be significant, the covariate with minimal $p$-value is selected for splitting. If no significant association is found, no split is conducted.  In the second step, the optimal split point is found by comparing two-sample linear statistics for all possible partitions for the split variable. The two-step approach of CIF leads to unbiased split variable selection if sampling is done without replacement and if a test statistic of quadratic form is used \cite{Strobl2007}. However, the selection bias is ignored by many researchers using random forests \cite{Strobl2014}. 

The disadvantage of standard CIF is that the association test for selecting the split variable is based on a linear rank statistic, while the optimal split is a dichotomous threshold-based split. An approach to avoid a change in the statistical approach for split variable and split point selection is to use maximally selected rank statistics \cite{Lausen1992,Lausen1994,Lausen2004} for the split point selection. With maximally selected rank statistics, the optimal split variable is determined using a statistical test for binary splits, which adjusts for the multiple testing of multiple possible split points. Thus, this approach naturally reduces split variable selection bias. For example, for dichotomous split variables, such as sex, no adjustments for multiple testing are required. In contrast, through the use of maximally selected rank statistics adjusted $p$-values are obtained for continuous covariates, such as gene expression levels. Since the exact distribution of the maximally selected rank statistic is generally unknown, $p$-values need to be approximated. In the \texttt{party} package \cite{Hothorn2014} these $p$-values can be approximated with a conditional Monte-Carlo approach, which is slow to compute for large datasets. We therefore have implemented random forests for survival endpoints using maximally selected rank statistics with several $p$-value approximations. We have investigated the split variable selection, prediction performance and runtimes of the different approaches in simulation studies and with real data.

In the next section, maximally selected rank statistics for survival outcome are introduced. Their implementation in a survival RF algorithm is described in Section~\ref{sec:implementation}. In Section~\ref{sec:selection}, the selection bias of different methods for the $p$-value approximation is investigated, and in Section~\ref{sec:simStudy2} the prediction performance of the new approach is compared with existing methods on simulated data. Finally, in Section~\ref{sec:realData}, prediction performance and runtime of the methods is evaluated on three real datasets on breast cancer, including a gene expression study and a genome-wide association study. 

\section{Maximally selected rank statistics for survival endpoints}\label{sec:maxstat}
In this section, we aim at finding an optimal cutpoint to split data in two groups using maximally selected rank statistics. A cutpoint is considered optimal if the separation of the survival curves in the two groups is maximized. 

We consider $n$ observations $(X_1,Y_1), \dots, (X_n,Y_n)$ and a continuous marginal distribution for $X$. $Y_i = (Z_i, \delta_i)$ is a right censored survival time with time $Z_i$ and censoring indicator $\delta_i$. We use log-rank scores $a_1,\dots,a_n$ \cite{Hothorn2003}
\begin{align*}
  a_i = a_i(\mathbf{Z}, \boldsymbol{\delta}) = \delta_i - \sum_{j=1}^{\gamma_i(\mathbf{Z})} \frac{\delta_j}{(n-\gamma_j(\mathbf{Z})+1)} \, ,
\end{align*}
where $\mathbf{Z}=(Z_1,\dots,Z_n)'$ and $\boldsymbol{\delta}=(\delta_1,\dots,\delta_n)'$ are the vectors of the survival time and censoring indicator, respectively, and $\gamma_j(\mathbf{Z}) = \sum_{i=1}^n \mathds{1}_{\{ Z_i \leq Z_j \}}$ is the number of observations with survival time up to $Z_j$. The linear rank statistic for a split by a cutpoint $\mu$ is the sum of all scores in the group with $X_i \leq \mu$
\begin{align*}\label{eq:teststat}
  S_{n\mu} = \sum_{i=1}^n \mathds{1}_{\{ X_i \leq \mu \}} a_i.
\end{align*}

The null hypothesis of no influence of a split with cutpoint $\mu$ on the distribution of $Y$ is $H_0\!: P(Y \leq y|X \leq \mu) = P(Y \leq y|X > \mu)$ for all $\mu$ and all $y$. Under the null hypothesis expectation and variance are given by \cite{Lausen1992}
\begin{displaymath}
  \mathbb{E}_{H_0}(S_{n\mu}|a,X) = m_\mu \bar{a} \quad \text{and} \quad
  \mathbb{V}ar_{H_0}(S_{n\mu}|a,X) = \frac{n}{n-1} \frac{m_\mu}{n} \frac{n_\mu}{n} \sum_{i=1}^{n} (a_i - \bar{a})^2 \, ,
\end{displaymath}
where $m_\mu = \sum_{i=1}^n \mathds{1}_{\{ X_i \leq \mu \}}$ and $n_\mu = \sum_{i=1}^n \mathds{1}_{\{ X_i > \mu \}} = n - m_\mu$ are the number of observations in the two groups, and $\bar{a} = \frac{1}{n} \sum_{i=1}^n a_i$ is the mean of the scores of all observations. To compare different splits we use the score test statistic
\begin{align*}
  T_{n\mu} = \frac{S_{n\mu} - \mathbb{E}_{H_0}(S_{n\mu}|a,X)}{\sqrt{\mathbb{V}ar_{H_0}(S_{n\mu}|a,X)}} \, .
\end{align*}

To maximize the separation of the data induced by a split, the cutpoint $\mu$ with maximal $T_{n\mu}$ is selected. To guarantee sufficiently large sample size in each group, the possible cutpoints $\mu$ are restricted to $\mu_1 \leq \mu \leq \mu_2$ with $\mu_1$ and $\mu_2$ corresponding to quantiles $\varepsilon_1, \varepsilon_2$ of the distribution of $X$. The maximally selected rank statistic is defined as 
\begin{align*}
  M_{n}(a,X,\varepsilon_1,\varepsilon_2) = \max_{\mu \in \left[\mu_1,\mu_2\right]} (|T_{n\mu}|) \, .
\end{align*}

In standard random forests the values of $T_{n\mu}$ would be compared not only between split points on the same variable but also between variables. As explained in Section~\ref{sec:introduction}, this would induce a bias if the number of possible splitting points differs between variables. Therefore, we obtain $p$-values for the maximally selected rank statistic $M_{n}(a,X,\varepsilon_1,\varepsilon_2)$ for each variable and compare the variables on the $p$-value scale. However, the exact distribution $P_{H_0}(M_{n}(a,X,\varepsilon_1,\varepsilon_2) \leq b)$, $b \in \mathbb{R}^+$
of the test statistic is generally unknown, and no exact $p$-values can be obtained. Several approximations have been proposed to the distribution of the exact test statistic under the null hypothesis, and asymptotic $p$-value approximations can be obtained as well. First, the exact distribution can be approximated using a standard permutation approach. Second, Lausen and Schumacher \cite{Lausen1992} proposed using the distribution of the supremum of the absolute value of a standardized Brownian bridge, to approximate the p-value by
  \begin{align*}
    P_{1}(b, \varepsilon_1,\varepsilon_2) = \frac{4 \phi(b)}{b} + \phi(b) \left( b - \frac{1}{b}\right) \log \left( \frac{\varepsilon_2 (1-\varepsilon_1)}{(1-\varepsilon_2) \varepsilon_1} \right),
  \end{align*}
  where $\phi(b)$ is the standard normal density. Lausen \textit{et al.} \cite{Lausen1994} provided an approximation using an improved Bonferroni inequality \cite{Worsley1982}
  \begin{align*}
    P_{2}(b, \varepsilon_1,\varepsilon_2) = 2 (1-\Phi(b)) + \sum_{i=1}^{k-1} D(l_i, l_{i+1}) \, ,
  \end{align*}
  where $l_1, \dots, l_k$ are the number of observations having values less or equal than the cutpoint for all $k$ possible cutpoints and
  \begin{align*}
    D(i,j) &= \frac{1}{\pi} \exp \left( -\frac{b^2}{2} \right) \left( t_{ij} - \left( \frac{b^2}{4} - 1 \right) \frac{t_{ij}^3}{6}\right)
  \end{align*}
  with
  \begin{align*}
    t_{ij} &= \sqrt{1-\frac{i(N-j)}{(N-i)j}} \, .
  \end{align*}
Lausen \textit{et al.} \cite{Lausen1994} recommended to use the minimum $p$-value of the two approximation methods because both are conservative. Hothorn and Lausen \cite{Hothorn2003} derived a lower bound of the distribution of the test statistic using the exact distribution of linear rank statistics by utilizing Streitberg and Röhmel's shift algorithm \cite{Streitberg1986}. Finally, the distribution of the test statistic can be approximated by a maximally selected Gauss statistic \cite{Hothorn2003} because linear rank statistics are asymptotically normally distributed under suitable regularity conditions \cite{Hajek1999}. The distribution of a maximally selected Gauss statistic can be computed using the algorithms by Genz \cite{Genz1992}.
  
\section{Survival forests with maximally selected rank statistics}\label{sec:implementation}
In this section, we propose a random forest algorithm for survival endpoints using maximally selected rank statistics (MSR-RF), compare it with standard random forests, RSF and CIF, describe the estimation of specific parameters and the implementation of the procedure using R and C++.

\subsection{Core elements of the algorithm}
The main elements of the algorithm are identical to those of the standard random forest algorithm. The number of bootstrap samples is fixed a priori. Next, a bootstrap sample is drawn from the data, and a tree is grown for each bootstrap sample. Alternatively, subsampling of the observations can be employed instead of bootstrapping. To induce randomness on the covariate level, a subset of covariates is randomly selected and made available for splitting at each node. The tree growing is stopped if a specific stop criterion is fulfilled. In the final step of the algorithm, results of the trees are aggregated, e.g., to estimate a prediction error. 

Our novel algorithm follows the basic concept of CIF. Thus, we propose to split nodes using a two-step procedure. First, for each candidate covariate, maximally selected rank statistics are computed, and the split point with the maximal standardized test statistic is selected. For each of these covariates, the $p$-value is obtained for the best split point under the null hypothesis of no association between the split point and the covariate. $p$-values are approximated or asymptotic $p$-values, estimated using one of the methods described in Section \ref{sec:maxstat}. The covariate with the smallest $p$-value is selected as splitting candidate, where $p$-values are adjusted using the Benjamini and Hochberg \cite{Benjamini1995} adjustment for the multiple testing of a total of \texttt{mtry} variables at a possible split point. If the adjusted $p$-value of the candidate covariate is not smaller than a pre-specified type I error level $\alpha$, no split is performed. Thus, only if the adjusted $p$-value of the candidate covariate is smaller than $\alpha$, the split is made. This part is one of the major differences between the proposed random forest algorithm and the standard random forest procedure. A major difference to CIF is in the second step: In CIF, the optimal split point is determined here. This simplifies in the proposed approach because the optimal split point is determined as a by-product in step $1$ of the procedure. It is the split point corresponding to the maximally selected rank statistic in step $1$.

In the original CIF approach \cite{Hothorn2006}, a linear rank test between log rank scores and the covariate is implemented in \texttt{party}. Thus, a linear association test is performed between a possibly continuous covariate and quantitative score values. Although a standard association test is performed in the first step, a standard binary split is done in the second step. With the MSR-RF procedure, the optimal binary split is determined as in standard random forests, but an adjustment for multiple possible splits is performed through the use of maximally selected rank statistics. As a result, no adjustments need to be made for dichotomous variables, such as sex. However, several adjustments are made for ordinal covariates, and adjustments play a major role for continuous covariates. An advantage of MSR-RF when compared with standard CIF is that binary splits are used for determining the best split and the optimal splitting variable. In this way, one procedure is used consistently in both steps of MSR-RF.

\subsection{Estimation}
In each terminal node $v$ of a tree, the survival function is estimated using the Kaplan-Meier estimator, utilizing only the observations from the same terminal node. For a single tree $b$, a prediction of the survival function $\hat{S}_b(t|\fe{x}_i)$ is made for a new observation $\fe{x}_i$ by dropping the observation down the tree. The prediction of the random forest $f$ with $B$ trees is obtained by averaging the predictions over all trees for each observation and time point: 
\begin{align}
  \hat{S}_f(t|\fe{x}_i) = \frac{1}{B} \sum_{b=1}^{B} \hat{S}_b(t|\fe{x}_i).
\end{align}

It is standard in random forests to use only the out-of-bag (OOB) samples for estimating the prediction error. 
We define $I_{i,b} = 1$ if $i$ is an OOB observation in tree $b$, and $0$, otherwise \cite{Ishwaran2008}. The OOB prediction for observation $i$ is then defined as
\begin{align}\label{eq:SfOOB}
\hat{S}_f^{OOB}(t|\fe{x}_i) = \sum\nolimits_{b=1}^{B} \big(I_{i,b} \cdot \hat{S}_b(t|\fe{x}_i) \big) \Big/ \sum\nolimits_{b=1}^{B} I_{i,b} \, .
\end{align} 

The prediction error is estimated with the Brier score (BS) for censored data \cite{Graf1999} using only the out-of-bag observations of each tree
\begin{align*}
  BS(t, \hat{S}_f^{OOB}) = \frac{1}{n} \sum_{i=1}^{n} \widehat{W}_i(t) \left( \mathds{1}_{\{ T_i > t \}} - \hat{S}_f^{OOB} (t|\fe{x}_i) \right)^2,
\end{align*}
where $T_i$ is the observed survival time for subject $i$. $\hat{S}_f^{OOB} (t|x_i)$ is the random forest estimator of the survival function, and $\widehat{W}_i(t)$ are weights using the inverse probability of censoring \cite{Mogensen2012}. For $t \in {[0, t^{\ast} ]}$, the integrated Brier score ($IBS$) is obtained as
\begin{align*}
  IBS(\hat{S}_f^{OOB}) = \int_{0}^{t^{\ast}} BS(t, \hat{S}_f^{OOB}) \, \mathrm{d}t .
\end{align*}

\subsection{Implementation in R}
The new method is implemented in the R \cite{RCoreTeam2014} package \texttt{maxstatRF} which is supplied with this article. To avoid unnecessary copying of data, the package is implemented using reference classes \cite{Wickham2014a}. Furthermore, the \texttt{parallel} package \cite{RCoreTeam2014} is used for parallel computation, and the package \texttt{ipred} \cite{Peters2013} is employed for calculating the BS. The \texttt{maxstat} package \cite{Hothorn2015} provides all necessary elements for the computation of maximally selected rank statistics. All $p$-value approximations described in Section~\ref{sec:maxstat} are available. The approximations are abbreviated as follows; \texttt{Lau92}:~Lausen and Schumacher \cite{Lausen1992}, \texttt{Lau94}:~Lausen \textit{et al.} \cite{Lausen1994}, \texttt{minLau}:~Minimum of Lausen and Schumacher \cite{Lausen1992} and Lausen \textit{et al.} \cite{Lausen1994}, \texttt{HL}:~Hothorn and Lausen \cite{Hothorn2003}, \texttt{condMC}:~Conditional Monte-Carlo, \texttt{exactGauss}:~Approximation by maximally selected Gauss statistic. These abbreviations will be used throughout the rest of this paper.

Tuning parameters in the \texttt{maxstatRF} package include standard random forest settings, such as the number of trees (\texttt{num\_trees}) in the random forest, the number of randomly selected variables considered for splitting in each node (\texttt{mtry}) and the minimal terminal node size (\texttt{min\_node\_size}). The parameter \texttt{minprop} corresponds to the lower quantile $\varepsilon_1$ of the covariate distribution (see Section \ref{sec:maxstat}). The upper quantile is set to $\varepsilon_2 = 1-\varepsilon_1$ by default. The significance level for splitting $\alpha$ and the $p$-value approximation method can be set as in the \texttt{maxstat} package. Finally, the sampling can be done with replacement (bootstrapping) or without replacement (subsampling). If the latter is used, the bootstrap sampling in the RF algorithm is replaced by subsampling with approximately $63.2\%$ of the observations. 

\subsection{Implementation in C++}
The software package \texttt{maxstatRF} is not optimized for computational efficiency. We also implemented the new method in the R package \texttt{ranger} \cite{Wright2016} to enable the efficient analysis of larger datasets, the growth of random forests with large \texttt{mtry} values, the tuning of the terminal node size or a large number of trees. The results from Section~\ref{sec:simStudy2} show that the $p$-value approximation by the minimum of the approximations by Lausen and Schumacher \cite{Lausen1992} and Lausen \textit{et al.} \cite{Lausen1994} leads to high prediction accuracy. Since other approximation methods show lower prediction accuracy or are computationally more intensive (Section~\ref{sec:selection}), only this approximation is implemented in \texttt{ranger}.

\section{Simulation study 1: split variable selection}\label{sec:selection}
Since approximations are used for the distribution of the maximally selected rank statistic, the split variable selection is potentially biased. Furthermore, the use of bootstrapping could induce a biased selection \cite{Strobl2007}. To assess both sources of potential bias we conducted a simulation study. Datasets were generated with a censored survival outcome and five covariates. The survival outcome was simulated by the minimum of a survival and censoring time, both assumed to be exponentially distributed with $\lambda = 0.5$ and $\lambda = 0.1$, respectively. Four out of five covariates were simulated with a discrete uniform distribution with a varying number of ordered categories: $x_{2}$, $x_{4}$, $x_{10}$ and $x_{20}$ with $2$, $4$, $10$ and $20$ categories, respectively. The $5^{th}$ covariate $x_n$ was simulated using the standard normal distribution to generate many unique values for the dataset. We simulated the null model \cite{Strobl2007} of totally uninformative predictor variables. In the ideal case, all covariates should thus be selected with equal frequencies. 

All simulations were run with $500$ trees, $\texttt{mtry} = 3$ and covariate quantiles of $\varepsilon_1 = 0.1$, thus $\varepsilon_2 = 0.9$. All six $p$-value approximation methods and splitting without adjustment for multiple testing were compared. Only the first split was investigated in each tree to avoid the influence of previous splits. For example, after one split on a dichotomous variable in a tree, this variable could not be selected again in that branch, and no unbiased split variable selection could be measured if further splits would be considered. We furthermore analyzed the effect of sampling with replacement (bootstrapping) and without replacement (subsampling). For subsampling a proportion of $63.2\%$ of the observations was used, which approximately equals the number of unique observation in a bootstrap sample. The number of replications was set to $1000$.

\begin{figure}
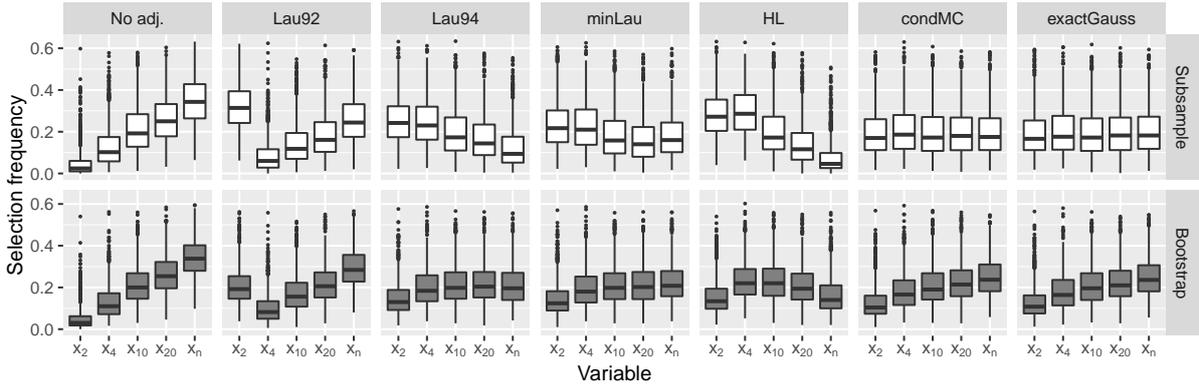

  \centering
  \includegraphics[width=1\linewidth]{{{rf_selection_n_100_minprop_0.1}}}
  \caption{Split variable selection for different $p$-value approximation methods for the null case of no association between covariates and survival time. Splitting without adjustment and adjustment by the $p$-value approximations of Lausen and Schumacher \cite{Lausen1992} (\texttt{Lau92}), Lausen \textit{et al.} \cite{Lausen1994} (\texttt{Lau94}), their minimum (\texttt{minLau}), the approximation by Hothorn and Lausen \cite{Hothorn2003} (\texttt{HL}), the permutation approach (\texttt{condMC}) and the maximally selected Gauss statistic (\texttt{exactGauss}) is compared. Subsampling, i.e., sampling without replacement, is displayed in the upper panel and bootstrapping, i.e., sampling with replacement, in the lower panel. The covariates $x_{2}$, $x_{4}$, $x_{10}$ and $x_{20}$ were simulated with $2$, $4$, $10$ and $20$ ordered categories, respectively, and the covariate $x_n$ is continuous.}
  \label{fig:rf_selection}
\end{figure}

Fig.~\ref{fig:rf_selection} displays the results for the different approximations at a sample size of $n = 100$. As expected, splitting without adjustment for multiple testing showed preferential splitting of covariates with a large number of possible split points for both bootstrapping and subsampling. The Lausen and Schumacher \cite{Lausen1992} (\texttt{Lau92}) approximation preferred variables with many categories for both bootstrapping and subsampling. However, the approximation proved to be conservative, since the variable with only one possible split point $x_2$, which does not require an adjustment for multiple testing, was often preferred. The Lausen \textit{et al.} \cite{Lausen1994} (\texttt{Lau94}) approximation was also conservative, since variables with few categories were preferred if subsampling was used. For bootstrapping, this bias was cancelled out by the bootstrapping bias and variables with many categories were preferred. However, in both cases, the bias was less pronounced than for the Lausen and Schumacher approximation. The combination \texttt{minLau} was similar to the approximation by Lausen et al. \cite{Lausen1994}, but in the case of subsampling the bias was slightly lower. As expected for an upper bound, the Hothorn and Lausen approximation \cite{Hothorn2003} (\texttt{HL}) showed selection bias towards covariates with few categories for both resampling schemes, and it was quite large in case of subsampling. Both the permutation approach (\texttt{condMC}) and the exact Gauss statistic (\texttt{exactGauss}) yielded unbiased splits in case of subsampling and preferential splitting of covariates with many categories in case of bootstrapping.

Supplementary Figures~$S1-S2$ display the results for varying sample sizes of $n = 25, 50, 100$ and $n = 200$ and lower covariate quantiles varying between $\varepsilon_1 = 0, 0.1, 0.25$ and $0.4$. In each case, the upper quantile was set to $\varepsilon_2 = 1-\varepsilon_1$. No results were obtained for the Hothorn and Lausen $p$-value approximation for $n = 200$ and bootstrapping because the approximation cannot handle sample sizes $> 171$. More specifically, the gamma function $\Gamma(\cdot)$ is used to compute the number of permutations, and $\Gamma(n)$ is out of double precision range on a $64$ bit computer system if $n > 171$. In the case without a covariate space restriction, i.e., $\varepsilon_1 = 0$ and $\varepsilon_2 = 1$, the Lausen and Schumacher \cite{Lausen1992} (\texttt{Lau92}) approximation generally preferred covariates with fewer categories. Otherwise, results were similar to those displayed in Figure~\ref{fig:rf_selection}. 

The split variable selection of RSF and CIF was also investigated. CIF was run in two modes, with the default setting and using maximally selected rank statistics with a conditional Monte-Carlo $p$-value approximation \cite{Hothorn2006}. The results are displayed in Figure~$S3$ (supplementary material). As expected, RSF was biased towards variables with many categories, while standard CIF was unbiased. CIF using maximally selected rank statistics was unbiased for subsampling and biased for bootstrapping.

\begin{figure}
  \centering
  \includegraphics[width=.85\linewidth]{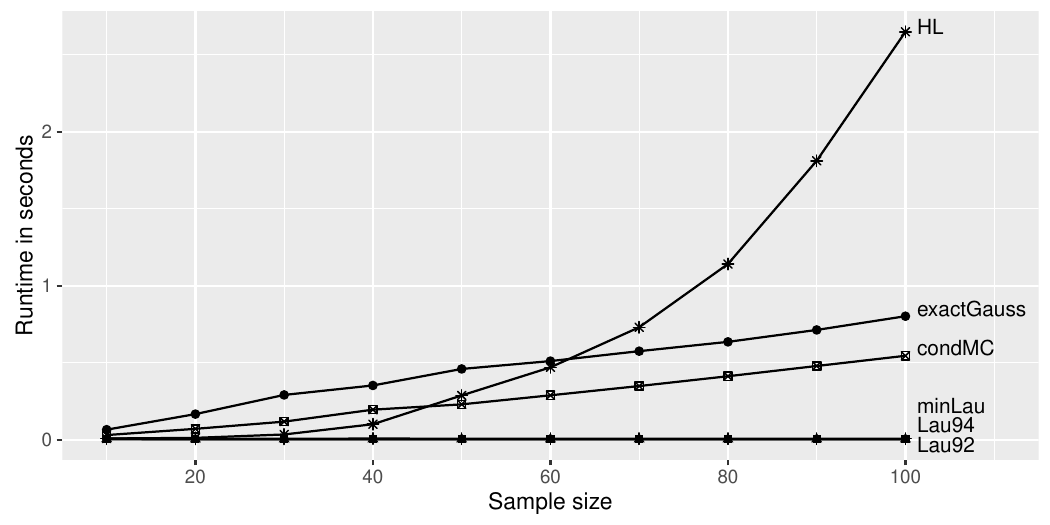}
  \caption{Runtime (in seconds) for varying sample size for different $p$-value approximations. Displayed runtimes correspond to the computation required in each node of a tree if $5$ covariates are considered and data is simulated as in Figure~\ref{fig:rf_selection}. The $p$-value approximations of Lausen and Schumacher \cite{Lausen1992} (\texttt{Lau92}), Lausen \textit{et al.} \cite{Lausen1994} (\texttt{Lau94}), their minimum (\texttt{minLau}), the approximation by Hothorn and Lausen \cite{Hothorn2003} (\texttt{HL}), the permutation Monte Carlo approach (\texttt{condMC}) and the maximally selected Gauss statistic (\texttt{exactGauss}) are compared. The runtimes of \texttt{Lau92}, \texttt{Lau94} are almost identical. The runtime of \texttt{minLau} is the sum of \texttt{Lau92} and \texttt{Lau94}.} 
  \label{fig:runtime}
\end{figure}

Since the $p$-value approximation is computed for each of the \texttt{mtry} splitting candidate covariates at every possible split in each tree, growing a forest with maximally selected rank statistics can be computationally intensive. We therefore compared the runtimes of the approximation methods for varying sample sizes. Data was simulated as in the previous simulation study with samples sizes between $10$ and $100$, and the $p$-values of the covariates were approximated using the different methods. Runtime results are summarized in Figure~\ref{fig:runtime} and Table~S1 (supplementary material). The runtimes of the approximations by Lausen and Schumacher \cite{Lausen1992} (\texttt{Lau92}) and Lausen \textit{et al.} \cite{Lausen1994} (\texttt{Lau94}) were approximately equal and smaller than for all other approximations. The runtime of the \texttt{minLau} method is the sum of the runtimes of \texttt{Lau92} and \texttt{Lau94}. The Hothorn and Lausen \cite{Hothorn2003} (\texttt{HL}) approximation was fast for small sample sizes, but runtime grew exponentially for larger sample sizes. Both the conditional Monte-Carlo approach \texttt{condMC} and the maximally selected Gauss statistic (\texttt{exactGauss}) scaled approximately linearly with the number of sample sizes, with a larger slope for \texttt{exactGauss}. 

\section{Simulation study 2: prediction performance}\label{sec:simStudy2}
To assess the prediction performance of the new method with the different $p$-value approximations, we conducted a second simulation study. We simulated datasets of two simulation scenarios and analyzed them with MSR-RF and all six $p$-value approximations, RSF and CIF in both variants as described in Section \ref{sec:selection}. The Cox model was included as a reference. 

\begin{figure}
  \centering
  \includegraphics[width=.45\linewidth]{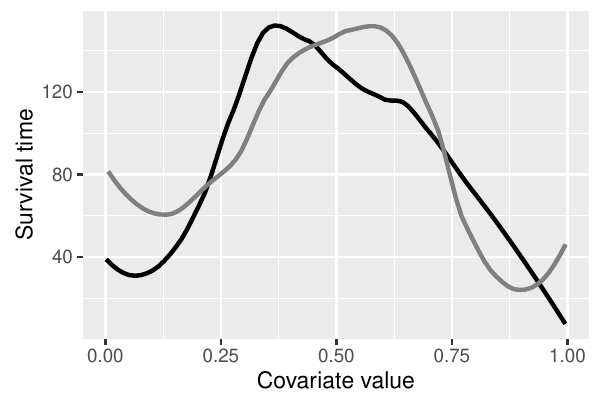}
  \caption{Effects of the non-linear covariates in simulation scenario $2$. The covariates were drawn from a continuous uniform distribution $U(0,1)$. The plotted data is LOESS smoothed with a span of $0.5$.} 
  \label{fig:nonlinear}
\end{figure}

The first simulation scenario was specifically designed to have few highly predictive variables with few categories and many non-predictive variables with many categories. The scenario was simulated with $200$ subjects, six dichotomous covariates with strong effect and $100$ normally distributed independent variables without effect. The second scenario consisted of $200$ subjects and $14$ covariates. Of these, two variables were continuously uniformly distributed with a strong non-linear effect as depicted in Figure~\ref{fig:nonlinear}. Two covariates were Bernoulli distributed and had a moderate linear effect, and $10$ covariates were normally distributed without effect. In both scenarios, the survival times and censoring times were simulated with exponential distributions. 

\begin{figure}
  \centering
  \includegraphics[width=1\linewidth]{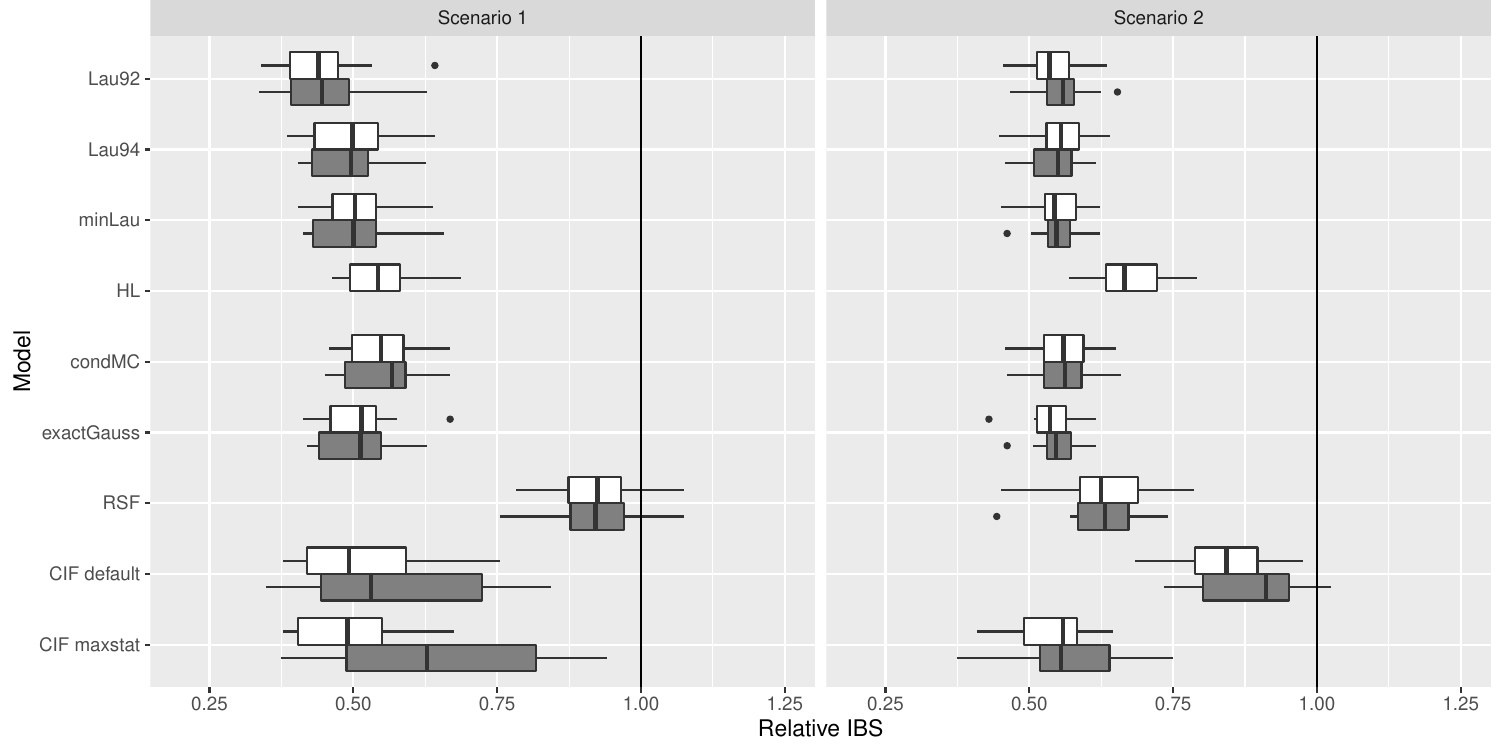}
  \caption{Prediction error for two simulation scenarios, different analysis methods and resampling schemes. The prediction errors are shown by boxplots (median, quartiles, largest non outliers) of integrated Brier scores (IBS) relative to the Cox model. Values left to the vertical line show results with lower IBS than the Cox model. White boxes correspond to subsampling, where $63.2\%$ of the samples were resampled without replacement. Grey boxes correspond to bootstrapping. The $p$-value approximations in MSR-RF are Lausen and Schumacher \cite{Lausen1992} (\texttt{Lau92}), Lausen \textit{et al.} \cite{Lausen1994} (\texttt{Lau94}), their minimum (\texttt{minLau}), the approximation by Hothorn and Lausen \cite{Hothorn2003} (\texttt{HL}), the permutation Monte Carlo approach (\texttt{condMC}) and the maximally selected Gauss statistic (\texttt{exactGauss}). In addition, random survival forests \cite{Ishwaran2008} (RSF), conditional inference forests \cite{Hothorn2006} with default settings (CIF default) and using maximally selected ranks statistics (CIF maxstat) are displayed. No results could be obtained for \texttt{HL} and bootstrapping because the approximation cannot handle sample sizes $> 171$; see Section~\ref{sec:selection} for details.} 
  \label{fig:performance_sim}
\end{figure}

To avoid overfitting, nested cross-validation was employed to assess prediction accuracy. We used ten folds for the outer cross-validation and five folds for the inner cross-validation. For each outer cross-validation fold, a model was built and tuned on the remaining nine subsamples, and the prediction error was estimated on the selected fold. For parameter tuning, the inner cross-validation was used. The integrated Brier score for censored data \cite[IBS]{Graf1999} was used for estimating the prediction errors, see Section~\ref{sec:implementation}. The R package \texttt{pec} \cite{Mogensen2012} was used for computing the IBS. The random forest analyses were run using $500$ trees and $\texttt{mtry} = p/2$, where $p$ is the number of covariates in the model. For the parameter tuning the number of trees was reduced to $50$ to reduce the computational effort. For MSR-RF, $\varepsilon_1 = 0.1$, and all six $p$-value approximation methods were used. All analyses were repeated for sampling with replacement (bootstrapping) and sampling without replacement (subsampling). 

As described in Section~\ref{sec:implementation}, the size of the trees in MSR-RF is determined by the significance level for splitting $\alpha$. For small values the tree growing is stopped early and larger values of $\alpha$ result in larger trees. In CIF, the tree size is also determined by a significance level, but a parameter $\texttt{mincriterion} = 1-\alpha$ is used. In RSF, no significance test is performed for splitting, and the tree size is controlled by limiting the minimal terminal node size with a parameter \texttt{nodesize}. The optimal size of trees heavily depends on the algorithm, the resampling method and the data. Therefore, \texttt{nodesize} was tuned in the nested cross-validation for each algorithm, dataset and subsampling combination. The parameters $\alpha = 0.1, 0.3, 0.5, 0.7, 0.9, 1$, $\texttt{nodesize} = 1, 3, 10, 25, 50$ and $\texttt{mincriterion} = 0, 0.1, 0.3, 0.5, 0.7, 0.9$ were evaluated for MSR-RF, RSF and CIF, respectively. For each combination, the parameter was selected that led to the lowest average IBS on all cross-validation folds. The optimal parameters are displayed in Table~S2 (supplementary material). 

The results of the benchmark experiments are shown in Figure~\ref{fig:performance_sim}. In the first simulation scenario (Scenario $1$), MSR-RF performed best with the Lausen and Schumacher \cite{Lausen1992} (\texttt{Lau92}) approximation, followed by Lausen \textit{et al.} \cite{Lausen1994} (\texttt{Lau94}), \texttt{minLau}, the approximation by the maximally selected Gauss statistic (\texttt{exactGauss}) and the two CIF approaches with subsampling. The permutation approach (\texttt{condMC}) and Hothorn and Lausen \cite{Hothorn2003} (\texttt{HL}) performed slightly worse. RSF performed similarly to the Cox model, but worse than the other methods. Systematic difference between subsampling and bootstrapping was observed for CIF only. In the second simulation scenario (Scenario $2$), MSR-RF performed best with all approximations but \texttt{HL}. CIF using maximally selected rank statistics performed similarly with a slightly higher variance. RSF performed slightly worse, while standard CIF and the Cox model performed substantially worse than the other models.

As expected, the RSF approach suffered from biased split variable selection if few highly predictive variables with few categories were combined with many non-predictive variables with many categories (Scenario $1$). However, if covariates with non-linear effects were included (Scenario $2$), the linear rank statistic in the CIF approach failed to select these covariates, which subsequently resulted in a reduced prediction performance. Except of the approximation by Hothorn and Lausen \cite{Hothorn2003} (\texttt{HL}), the approximations for MSR-RF were comparable in terms of prediction performance. As expected, methods biased towards covariates with few categories performed better in Scenario $1$.

\section{Real data applications: survival prediction of breast cancer patients}\label{sec:realData}
To demonstrate the efficiency of the proposed approach and to compare the prediction performance with other methods, we analyzed three real datasets on breast cancer with MSR-RF, RSF and CIF in both variants as described in Section \ref{sec:selection}. Furthermore, we measured the runtime required for analyzing these datasets. 

The \texttt{GBSG2} dataset \cite{Schumacher1994} consists of eight clinical covariates for $686$ women from the German breast cancer study group $2$, a controlled clinical trial on the treatment of lymph node positive breast cancer for the endpoint recurrence-free survival. The covariates are hormonal therapy (yes or no), age, menopausal status, tumor size and grade, number of positive lymph nodes, progesterone receptor and estrogen receptor. The dataset is freely available in the \texttt{pec} R package \cite{Mogensen2012}. 

The \texttt{nki} dataset \cite{VanDeVijver2002, vantVeer2002} includes gene expression measurements of $337$ lymph node positive breast cancer patients. Seven clinical risk factors and $24,481$ genes are included. We analyzed the endpoint metastasis-free survival. The dataset is freely available in the Bioconductor \cite{Gentleman2004} package \texttt{breastCancerNKI} \cite{Schroeder2011}.

The \texttt{success} study is a randomized phase III trial on the treatment of breast cancer, and it includes a genome-wide association study (GWAS). Originally, the dataset included $3322$ patients, $17$ clinical variables and $693,543$ single nucleotide polymorphisms (SNPs). We performed a standard quality control and linkage disequilibrium pruning ($r^2 > 0.7$), which reduced the dataset to $2781$ individuals and $331,195$ SNPs. Finally, to reduce the computational burden and eliminate missing data, we excluded all SNPs with a call fraction below $100$\%, keeping $151,346$ SNPs. We analyzed the endpoint relapse-free survival. This dataset is available at dbGaP and has ID \texttt{phs000547.v1.p1}.

Prediction accuracy was estimated from nested cross-validation using the IBS as in Section~\ref{sec:simStudy2}. The random forest analyses were run using $500$ trees and $\texttt{mtry} = p/2$, where $p$ is the number of covariates in the model. For MSR-RF, we used $\varepsilon_1 = 0.1$. Because of the size of the datasets, only the minimum of the $p$-value approximations by Lausen and Schumacher \cite{Lausen1992} and Lausen \textit{et al.} \cite{Lausen1994} was used. All analyses were repeated for sampling with replacement (bootstrapping) and sampling without replacement (subsampling). The R packages \texttt{randomForestSRC} \cite{Ishwaran2014}, \texttt{party} \cite{Hothorn2014} and \texttt{ranger} \cite{Wright2016} were used for RSF, CIF and MSR-RF, respectively. The parameters to control the tree size were tuned as in Section~\ref{sec:simStudy2}. The number of trees in the parameter tuning was reduced to $50$ because of runtime restrictions. Tuning results are displayed in Table~S2 in the supplementary material. To measure the runtime on complete datasets, it was measured in an additional run without cross-validation. Runtime was measured using the tuned parameter values.

\begin{figure}
  \centering
  \includegraphics[width=1\linewidth]{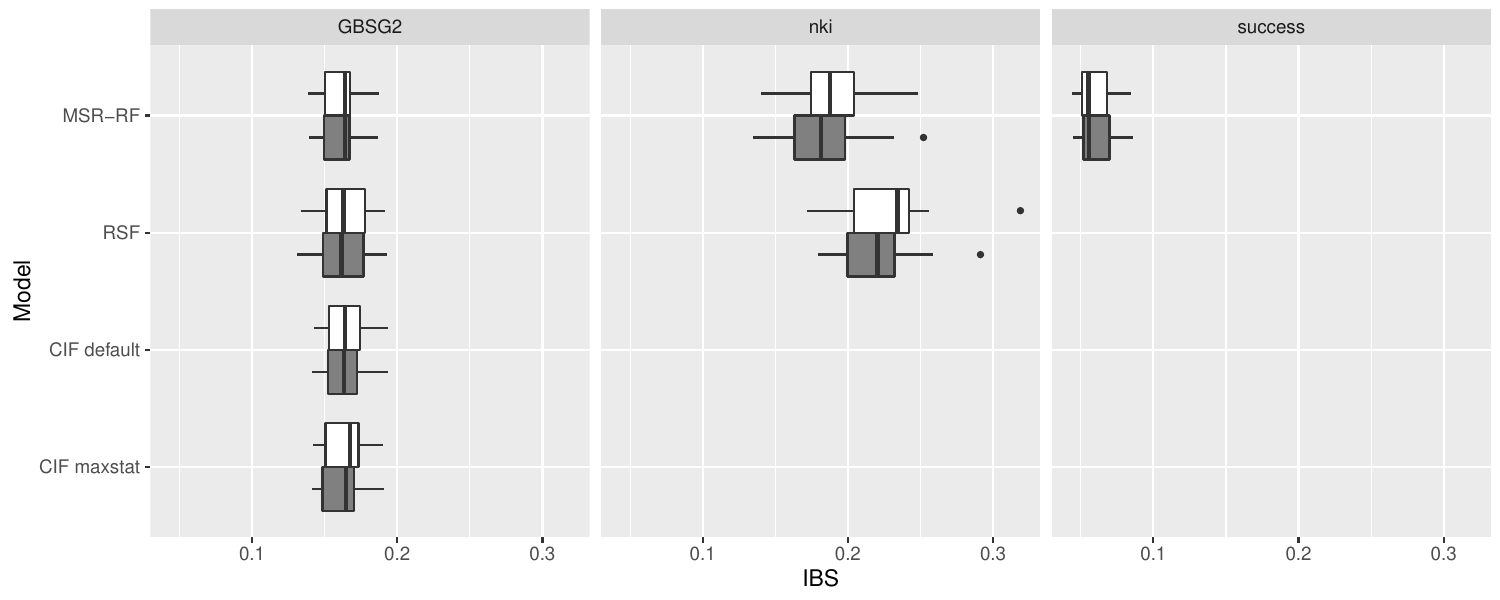}
  \caption{Prediction error for three breast cancer datasets, different analysis methods and resampling schemes. The prediction errors are shown by boxplots (median, quartiles, largest non outliers) of integrated Brier scores (IBS). White boxes correspond to subsampling, where $63.2\%$ of the samples were resampled without replacement. Grey boxes correspond to bootstrapping, where sampling was done with replacement. The maximally selected rank statistics random forest (MSR-RF) results were obtained using the R package \texttt{ranger} \cite{Wright2016} for the minimum of the $p$-value approximations by Lausen and Schumacher \cite{Lausen1992} and Lausen \textit{et al.} \cite{Lausen1994}. Random survival forests \cite[RSF]{Ishwaran2008} and conditional inference forests \cite{Hothorn2006} with default settings (CIF default) and using maximally selected rank statistics (CIF maxstat) are additionally included.} 
  \label{fig:performance_real}
\end{figure}

Prediction performance is shown in Figure~\ref{fig:performance_real}, runtime in Table~\ref{tab:runtime}. In the \texttt{GBSG2} dataset, all methods performed similarly. MSR-RF was fastest in computation, followed by RSF and standard CIF. CIF using maximally selected rank statistics took over 3 hours, while the other methods took only a few seconds to analyze the \texttt{GBSG2} dataset. In the \texttt{nki} dataset, MSR-RF outperformed RSF. Both variants of CIF did not complete the analysis. Specifically, we stopped the computation after four days without any results. The computation time of MSR-RF was substantially lower than the computation time of RSF. Finally, for the \texttt{success} GWAS dataset, only MSR-RF completed the analysis. Both CIF and RSF aborted with a memory error, even if run on a high performance computing node with $192$ GB system memory. For all three datasets, no substantial differences were observed between subsampling and bootstrapping. 

\begin{table}
\caption{Runtime (in minutes) to grow a random forest with $500$ trees using the new maximally selected rank statistics random forest (MSR-RF) approach, random survival forests \cite[RSF]{Ishwaran2008} and conditional inference forests \cite{Hothorn2006} with default settings (CIF default) and using maximally selected ranks statistics (CIF maxstat). The MSR-RF results were obtained using the R package \texttt{ranger} \cite{Wright2016} and the minimum of the $p$-value approximations by Lausen and Schumacher \cite{Lausen1992} and Lausen \textit{et al.} \cite{Lausen1994}. All results are for sampling with replacement and the tuned parameters. For MSR-RF and RSF, multithreading was used with $12$ threads. \textsuperscript{*}Stopped after $4$ days of computation. \textsuperscript{\textdagger}Memory error.}
\label{tab:runtime}
\centering
\begin{tabular}{lrrr}
   \toprule
  Method & \multicolumn{3}{c}{Runtime in minutes} \\
  & \texttt{GBSG2} & \texttt{nki} & \texttt{success} \\ 
  \midrule  
   MSR-RF & $0.01$ & $1.73$ & $41.69$ \\
   RSF & $0.03$ & $58.64$ & NA\textsuperscript{\textdagger} \\ 
   CIF default & $0.04$ & $>5760$\textsuperscript{*} & NA\textsuperscript{\textdagger} \\ 
   CIF maxstat & $225.11$ & $>5760$\textsuperscript{*} & NA\textsuperscript{\textdagger} \\ 
  \midrule
\end{tabular}
\end{table}

\section{Discussion}\label{sec:discussion}
Random forests for survival data with maximally selected rank statistics (MSR-RF) reduce split variable selection bias and are able to deal with non-linearity in the covariates when compared with RSF and CIF. RSF is based on the standard random forest algorithm, and it showed the expected split variable selection bias \cite{Strobl2007} in Monte-Carlo simulation studies. Specifically, MSR-RF and CIF both outperformed RSF in the simulation scenarios with mixed covariate types because RSF favors covariates with many possible split points over covariates with fewer possible split points. In a simulation study with non-linear covariate effects, we observed a better performance of MSR-RF and RSF when compared with standard CIF. This finding can be explained by the fact that the standard CIF utilizes a linear rank statistic, which cannot cope with non-linearities. However, we stress that none of the random forest methods failed completely on any of the considered simulated or real data sets. In fact, the random forest methods performed equally well or even better than the standard Cox model in our simulations.

With real data examples, we have demonstrated the computational efficiency of the proposed MSR-RF approach. In a dataset with few clinical covariates it performed as well as CIF and RSF, and the runtime was lowest. In a gene expression dataset, CIF was unable to complete the analysis, while MSR-RF had a lower prediction error compared to RSF. Since this dataset is a mixture of strong clinical covariates and many noisy gene expression values, this confirmed the observations of the simulation studies. In addition, MSR-RF was substantially faster than RSF in this example. Finally, in a genome-wide association study, the analysis could be completed with the fast implementation of MSR-RF only. 

However, the MSR-RF version implemented in \texttt{ranger} is a compromise between the reduction in split variable selection bias and computational efficiency. In fact, in our comparison of six approaches for approximating the maximally selected rank statistic, only the Monte-Carlo permutation approach and the maximally selected Gauss statistic achieved unbiased split variable selections. However, both approaches are computationally intensive which restricts their usage to small random forests. Of the remaining methods, the minimum of the approximations by Lausen \textit{et al.} \cite{Lausen1994} and Lausen and Schumacher \cite{Lausen1992} achieved the best results. The method performed equally well for small sample sizes. Since the minimum approach is fast to compute, we consider this approach as the best compromise in the tradeoff between split variable selection bias and runtime for larger random forests.

MSR-RF are just one special case of the general approach to maximally selected statistics \cite{Hothorn2008}. It is easily extendable to continuous outcome by setting the scores equal to the ranks \cite{Lausen1992}. Furthermore, the approach can be extended to categorical endpoints by using a maximally selected $\chi^2$ statistic \cite{Miller1982}. For the binary case, exact methods to obtain $p$-values are known \cite{Koziol1991,Boulesteix2006,Boulesteix2006a}, and for the categorical case approximations have already been proposed \cite{Betensky1999}. An alternative for the binary case is to code the outcome as $0$ and $1$ and use maximally selected rank statistics \cite{Potapov2012}. 

The restriction of possible split points to an interval $[\mu_1, \mu_2]$, corresponding to quantiles $\varepsilon_1, \varepsilon_2$ of the distribution of the covariates, suppresses extreme end-cut splits, to provide sufficient sample sizes in both groups for asymptotic properties to hold for the $p$-value approximation \cite{Lausen1992}. However, recent findings \cite{Ishwaran2015} indicate that the ability to produce end-cut splits is desirable in random forests and our results in Section~\ref{sec:selection} suggest that the approximations work well for small sample sizes. Therefore, one could use a smaller value for the parameter \texttt{minprop} or even set it to $0$, not restricting the possible split points at all and possibly increasing the prediction accuracy in high-noise scenarios.

In Section~\ref{sec:maxstat}, we used a continuous marginal distribution for the covariates. This assumption reduces the number of possible cut points in one covariate, and it prevents from the analysis of unordered categorical covariates. Allowing splits on all possible $2$-partitions in the covariate space could solve this. However, this would lead to both an exponential increase in computation time and an increase in the multiple testing problem. As an alternative, the covariate categories could be ordered before splitting as proposed by Breiman \textit{et al.} \cite{Breiman1984}. For non-censored continuous outcomes in standard RF, ordering by the average outcome per category results in the same optimal split points, as if all $2$-partitions would be considered \cite{Fisher1958}. This could be extended to survival outcomes, e.g., by using median survival times or appropriate quantiles.

In this work, we used the Benjamini and Hochberg \cite{Benjamini1995} approach to adjust for multiple testing and to decide whether tree growing should be stopped. Other approaches could, of course, be used to adjust for the multiple testing of \texttt{mtry} variables at a node. A natural choice would be Hommel's procedure \cite{Hommel1988} which relies on the positive stochastic dependence of covariates \cite{Goeman2014}. However, we consider these alternative approaches to be computationally more demanding. Computational efficiency has therefore led to the choice of the Benjamini Hochberg procedure. The package \texttt{ranger} is quite flexible, and the interested user can easily add his/her preferred procedure.

Our results suggest that the choice of the resampling scheme, i.e., subsampling or bootstrapping, has no general impact on the prediction performance. However, the tuned parameters were different, indicating that for the same parameter combination the prediction performance of the methods differs. Since unbiased split variable selection is only possible with subsampling (Section~\ref{sec:selection}) and most of the theoretical results on random forests are obtained in this case \cite{Biau2015,Wager2014b}, we generally recommend to use subsampling.

Random forests with maximally selected rank statistics (MSR-RF) are a useful alternative to other random forest approaches for survival analysis. By the use of a two-step procedure, bias in split variable selection is reduced considerably. The method outperforms the Cox model on simulated datasets. Compared to other random forest approaches for survival analysis, the method performs better than random survival forests if informative dichotomous variables are combined with uninformative variables with more categories and better than conditional inference forests if non-linear covariate effects are included. In none of the analyzed scenarios the method is outperformed by another approach. With simple $p$-value approximations the method is computationally efficient and capable of analyzing very large datasets. 

\section*{Acknowledgements}
This work was financially supported by the European Union Seventh Framework Programme (FP7/2007-2013) under grant agreement No.\ HEALTH-F2-2011-278913 (BiomarCaRE).

\bibliography{maxstatRF.bib}
\bibliographystyle{wileyj}
\end{document}